\pdfoutput=1

\documentclass[11pt]{article}

\usepackage[final]{acl}

\usepackage{times}
\usepackage{latexsym}

\usepackage{amsmath,amssymb,amsfonts,amsthm}
\usepackage{mathtools}
\usepackage{bm}
\usepackage{subcaption}
\usepackage{cleveref}
\usepackage{xspace}
\usepackage{wrapfig, framed, caption}
\usepackage{makecell}

\usepackage{extarrows}

\usepackage{enumitem,kantlipsum}
\usepackage{thm-restate}
\usepackage{algorithm,algorithmicx,algpseudocode}

\newtheorem{theorem}{Theorem}[section]

\theoremstyle{definition}
\newtheorem{definition}[theorem]{Definition}

\theoremstyle{remark}

\usepackage[T1]{fontenc}

\usepackage[utf8]{inputenc}

\usepackage{microtype}

\usepackage{inconsolata}

\usepackage{graphicx}
\usepackage{forest}
\usepackage{xcolor}
\usepackage{booktabs} 
\usepackage{rotating}

\title{Large Language Models and Causal Inference in Collaboration: A Survey}

\author{
 \textbf{Xiaoyu Liu\textsuperscript{1,*}},
 \textbf{Paiheng Xu\textsuperscript{1,*}},
 \textbf{Junda Wu\textsuperscript{2}},
 \textbf{Jiaxin Yuan\textsuperscript{1}},
 \textbf{Yifan Yang\textsuperscript{1}},
\\
 \textbf{Yuhang Zhou\textsuperscript{1}},
 \textbf{Fuxiao Liu\textsuperscript{1}},
 \textbf{Tianrui Guan \textsuperscript{1}},
 \textbf{Haoliang Wang\textsuperscript{3}},
 \textbf{Tong Yu\textsuperscript{3}},
\\
 \textbf{Julian McAuley\textsuperscript{2}},
 \textbf{Wei Ai\textsuperscript{1}},
 \textbf{Furong Huang\textsuperscript{1}}
\\
\\
 \textsuperscript{1}University of Maryland, College Park,
 \textsuperscript{2}University of California San Diego,
 \textsuperscript{3}Adobe Research
\\
 \small{
   \textbf{Correspondence:} \href{furongh@umd.edu}{furongh@umd.edu}.
   * denotes equal contribution.
 }
}

\begin{document}
\maketitle
\begin{abstract}
Causal inference has demonstrated significant potential to enhance Natural Language Processing (NLP) models in areas such as predictive accuracy, fairness, robustness, and explainability by capturing causal relationships among variables. The rise of generative Large Language Models (LLMs) has greatly impacted various language processing tasks. This survey focuses on research that evaluates or improves LLMs from a causal view in the following areas: reasoning capacity, fairness and safety issues, explainability, and handling multimodality. Meanwhile, LLMs can assist in causal inference tasks, such as causal relationship discovery and causal effect estimation, by leveraging their generation ability and knowledge learned during pre-training. This review explores the interplay between causal inference frameworks and LLMs from both perspectives, emphasizing their collective potential to further the development of more advanced and robust artificial intelligence systems.
\end{abstract}

\section{Introduction}
Recently Large Language Models (LLMs) have showcased remarkable versatility across a spectrum of critical tasks. 
Moreover, there has been a recent expansion into multi-modal variants, such as Large Vision Language Models (VLMs) or Multi-modal Large Language Models (MLLMs), which broaden their input/output capabilities to encompass various modalities. This evolution has significantly enhanced both the potential and range of applications of these models. 
In this survey, our primary focus is on Transformer-based LLMs and LVLMs. 

The capability of LLMs is fundamentally rooted in their inference abilities, which dictates their proficiency in comprehending, processing, and providing solutions to various inquiries, as well as their adaptability to societally impactful domains~\cite{zhao2023survey}. Consequently, extensive research efforts have been dedicated to measuring and enhancing these capabilities, ranging from assessing the reasoning abilities of LLMs to scrutinizing their decision-making processes and addressing challenges such as concept alignment across different modalities and mitigating hallucination. Meanwhile, causal inference has shown great potential in improving predictive accuracy, fairness, robustness, and explainability of Natural Language Processing (NLP) models~\cite{feder-etal-2022-causal}, 
With LLMs revolutionizing various language processing tasks, there is a growing trend in applying causal inference to address LLM-related challenges and enhance their functionality. 
This survey outlines causal methodologies and their implementation in LLMs, emphasizing their role in enriching our comprehension and application of language models.

Moreover, this survey also aims to explore how LLMs can help with the causal inference framework. 
Causal inference is formally defined as an intellectual discipline that considers the assumptions, study designs, and estimation strategies that allow researchers to draw causal conclusions based on data \cite{pearl2009causality}. 
Causal inference has three main origins: potential outcomes, graphs, and structural equations, each serving unique purposes~\cite{zeng2022survey}.
In this survey, we mainly discuss Pearl's formulation of causal graphs \cite{pearl1998graphical}, which formalized causal graphical models for presenting conditional independence among random variables using directed acyclic graphs (DAGs). 

%

\begin{figure*}[!t]
  \centering
  \resizebox{1\textwidth}{!}{%
  \begin{forest}
  for tree={
    draw=red,
    align=center,
    anchor=west,
    child anchor=west,
    grow'=0,  
    rounded corners,
    font=\scriptsize
  }
  [Causal and Large Language Models (LLMs), rotate=90, anchor=south, text width=5.5cm
    [Causal for LLMs, 
      [Reasoning Capacity
        [Model Understanding
            [\citet{zevcevic2023causal, romanou2023crab, kim2023can, li2023relation} \\
            \citet{abdali2023extracting, tang2023towards, zhang2023mitigating,chen2024causal}, fill=yellow!20]
        ]
        [Commonsense Reasoning
            [\citet{gao2023chatgpt, willig2022probing, zhang2022rock}\\
            \citet{zheng2023preserving,lu2022neuro,chen2023learning}, fill=yellow!20]
        ]
        [Counterfactual Reasoning
            [\citet{betti2023relevance, miao2023generating, li2023large, liu2023magic}, fill=yellow!20]
        ]
        ]
      [Fairness and Bias
        [\citet{meade2021empirical, achiam2023gpt, stanovsky-etal-2019-evaluating, ding2022word, zhou2023causal, wang2023causal} \\
        \citet{madhavan2023cfl, jenny2023navigating,xia2024aligning,wu2024decot,gallegos2024bias}, fill=yellow!20]
      ]
      [Safety
        [\citet{cao2022can, bao2021defending, zhao2023causality, zhao2022certified, meng2021self, liu2023hallusionbench}, fill=yellow!20]
      ]
      [Explanation
        [\citet{hou2023towards, belrose2023eliciting, gurnee2023finding, zhao2023explainability,liu2023aligning,vig2020investigating}, fill=yellow!20]
      ]
      [Multi-modality
        [\citet{pawlowski2023answering, ko2023large, li2023image, su2023language, tai2023link, zhao2023causal,chen-etal-2024-quantifying}, fill=yellow!20]
      ]
      [Benchmark
        [\citet{xie2023echo, zhang2023causal, huang2023clomo, yu2023ifqa, jin2023cladder, nie2023moca} \\
        \citet{jin2023can,su2023language, li2023image, chen2023models, gao2023chatgpt,chen2024causal,chen-etal-2024-quantifying}, fill=yellow!20]
      ]
    ]
    [LLMs for Causal
      [Treatment Effect Estimation
        [\citet{chen2023disco, feder2023causal, zhang2023towards}, fill=yellow!20]
      ]
      [Causal Relationships Discovery
        [\citet{kiciman2023causal, naik2023applying, Antonucci2023ZeroshotCG, long2023can, vashishtha2023causal} \\
        \citet{arsenyan2023large, ban2023causal, tu2023causal,ban2023query, ban2023causal} \\
        \citet{joshi2024llms,feng2024pre,li2024realtcd,khatibi2024alcm,wan2024bridging}, fill=yellow!20]
      ]
    ]
  ]
\end{forest}
  }
    \caption{An overview of the interplay between causal inference frameworks and LLMs.}
    \label{fig:overview}
\end{figure*}
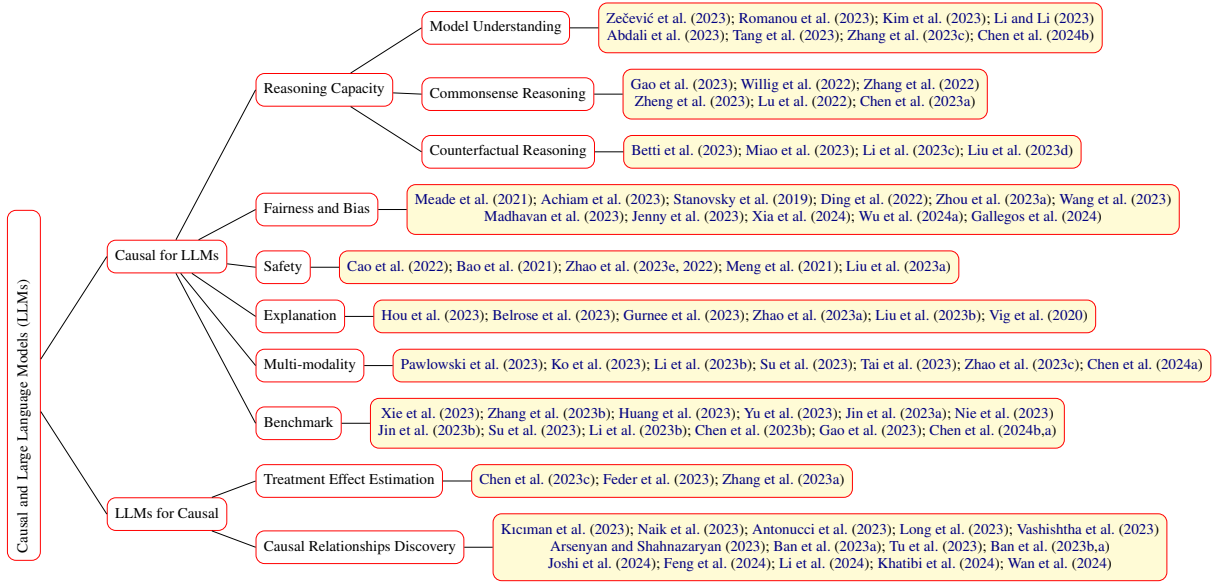

We summarize how LLMs can help causal inference in its two important components, i.e., causal relationship discovery and treatment effect estimation.
Estimating causal effects between variables requires assumptions about their relationships with other variables, traditionally provided by experts
LLMs, leveraging pre-trained knowledge, can assist in identifying these relationships and enhance causal discovery methods~\cite{zanga2022survey} for more reliable outcomes.
Additionally, estimating treatment effects is often hindered by the absence of counterfactual data. By utilizing LLMs' strong generative abilities, researchers have developed various ways to generate high-quality counterfactuals to enable treatment effect estimation. Figure \ref{fig:overview} shows an overview of selected topics about the interplay between causal inference frameworks and LLMs from both perspectives.

\section{Causal Inference and Large Language Models}


In this section, we provide a brief introduction to both large language models (LLMs) and causal inference, laying the groundwork for understanding their interaction. We begin by outlining the core principles and recent advancements in LLMs, which have reshaped natural language processing by enabling sophisticated language generation and comprehension. Following this, we introduce the fundamental concepts of causal inference, emphasizing its role in discerning cause-and-effect relationships within data. By establishing a foundational understanding of these two fields, we aim to clarify how their integration can lead to significant advancements, enhancing both the reasoning capabilities of LLMs and the development of causal inference methodologies.

\subsection{Evolution of the Large Language Models}
\label{sec:LLM}
Large Language Models (LLMs) have transformed the way we interact with and process language, opening up new possibilities for natural language understanding, generation, and communication \cite{zhao2023survey}. 
In this paper, we mainly focus on Transformer \cite{vaswani2017attention} based LLMs, and we provide an overview of their recent progress in this section. 

The introduction of the highly parallelizable Transformer architecture \cite{vaswani2017attention}, which leverages self-attention mechanisms, has led to the development of a range of Pre-trained Language Models (PLMs) with varying architectures and pre-training strategies, e.g., BERT~\cite{devlin-etal-2019-bert}, GPT2~\cite{radford2019language}, and so on \cite{radford2018improving, Liu2019RoBERTaAR, lewis-etal-2020-bart}. 
These models largely improved performances for NLP tasks with a learning paradigm of general-purpose pre-training and task-specific fine-tuning.
Researchers then find that scaling the size of the model or pre-training data often leads to further improved model capacity on downstream tasks \cite{kaplan2020scaling,wei2022emergent,schaeffer2024emergent}.
These larger-scaled PLMs are referred to as large language models (LLMs) \cite{brown2020language,Chowdhery2022PaLMSL,team2023gemini,achiam2023gpt,bai2023qwen,touvron2023llama,dubey2024llama}. 

These LLMs demonstrated strong capacities for language processing and solving complex tasks through text generation. 
We briefly introduce three common strategies that are used to further enhance their performance: In-context Learning (ICL) \cite{brown2020language}, which allows LLMs with natural language instruction and/or several task demonstrations (i.e., input-output pairs); 
Chain-of-Through (CoT) prompting \cite{wei2024cot}, which includes intermediate reasoning steps into prompts to guide the model toward the final answer; 
and Instruction tuning \cite{sanhmultitask,ouyang2022training,thoppilan2022lamda}, which refers to fine-tuning the LLMs with a mixture of multi-task datasets formatted as natural language instructions.
Additionally, researchers have extended these LLMs to handle images as input \cite{team2023gemini,achiam2023gpt}, leading to the development of Large Vision Language Models (VLMs) or Multi-modal Large Language Models (MLLMs).
In this survey, we primarily focus on works that adopt a causal perspective to improve LLMs in their generative capacities.

\subsection{Introduction of Causal Inference}

In this section, we present the background knowledge of causal inference, including task descriptions, basic concepts and notations, and general solutions. More details can be found in \Cref{sec:causal-basics}.

Causal inference aims to estimate the causal relationship among variables. The variables of interest are referred to as \textit{treatment}, naturally, the effects of treatments are referred to as \textit{treatment effects}. 
Ideally, the treatment effect can be measured as follows: applying different treatments to the same cohort, and then the difference in the effect is the treatment effect. However, it is impracticable for perfectly controlled experiments in many cases, requiring estimation of treatment effect from observational data. 
One of the most influential frameworks in identifying and quantifying causal effects in observational data is the \textbf{potential outcomes framework} \cite{rubin1974estimating}. The potential outcomes approach associates causality with manipulation applied to \textit{units}, and compares causal effects of different treatments via their corresponding potential outcomes. 

\begin{definition}[Binary Average Treatment Effect(ATE)] Suppose we want to measure the treatment effect of a treatment $T=1$. Then the average treatment effect is defined as:
\begin{equation}
    \label{def:ate}
    \mathbb{E}[Y(T=1) - Y(T=0)]
\end{equation}
where $Y(T=1)$ and $Y(T=0)$ denote the potential treated and control outcome of the whole population respectively.
\end{definition}

The potential outcome framework is powerful in recovering the effect of causes. In a potential outcome framework, causal effects are answered by specific manipulation of treatments. However, when it comes to identifying the causal pathway or visualizing causal networks, the potential outcome model has its limitations. In the front of the challenge, \textbf{causal graphical models} utilize directed edges to represent causalities and encode conditional independence among variables in the graphs. One of the most widely-spread formulations is the \textbf{Structural Equation Model} \cite{wright1934method, pearl1998graphical}, where linear structural equation models are used to present causal relationships by directed edges, which differentiate correlation from causation when the graph structure is given. The linearity assumption was later relaxed by \cite{pearl1998graphical} and it formalized causal graphical models for presenting causal relations using Directed Acyclic Graphs (DAGs).

Specifically, consider the random variable $\mathbf{X} \in \mathcal{R}^{D \times N} = [X_1, X_2, ..., X_N]$, where $D$ represents the dimensionality of each variable, and $N$ is the number of variables, the linear SEM consists of a set of equations of the form:
\begin{equation}
    \label{def:linear-sem}
    X_i = \beta_{0i} + \sum_{j \in pa(X_i)} \beta_{ji}X_j + \epsilon_i, \quad i = 1,2,3,..., N
\end{equation}
where $pa(X_i)$ denotes the set of variables that are direct parents of $X_i$. $\epsilon_1, \epsilon_2, ..., \epsilon_N$ are mutually independent noise terms with zero mean, $\beta_{ji}$ are coefficients that quantify the causal effect of $X_j$ on $X_i$.

While the non-parametric SEM takes the form:
\begin{equation}
    \label{def:non-linear-sem}
    X_i = f_i(\mathbf{X}_{pa(i)}, \epsilon_i), \quad i = 1,2,3,..., N
\end{equation}

The random variables $\mathbf{X}$ that satisfies the model structure of the form in \Cref{def:linear-sem} or \Cref{def:non-linear-sem} can be represented by a directed acyclic graph (DAG) $G = (V, E)$, where $V$ is the set of associated vertices, each corresponding to one of a variable of interest $X_i$, and $E$ is the corresponding edge set.
With pre-specified DAG and assumptions on the latent variables, the coefficients between the latent variables are identifiable \cite{kuroki2014measurement}.
Next, we show a comprehensive survey of how existing causal frameworks help challenges in LLMs.

\section{Causal Inference for Large Language Models}
\label{sec:causal_for_llm}

LLMs can significantly benefit from causal inference as it enhances their ability to understand and reason about cause-and-effect relationships within data. In this section, we review how LLMs can benefit from a causal lens in various capacities. We show an overview in Figure \ref{fig:overview}.

\subsection{Reasoning Capacity}
\label{sec:reasoning_capacity}


\subsubsection{Model Understanding}
\label{subsec:understanding}
LLMs have demonstrated many emerging abilities in language generation and certain reasoning tasks \cite{bubeck2023sparks,kiciman2023causal}.
As the reasoning process is often associated with causal factors, it is logical to first understand and evaluate LLMs' reasoning ability from a causal lens.
\citet{zevcevic2023causal} 
argued LLMs are not causal and 
hypothesized that LLMs are simply trained on the data, in which causal knowledge is embedded. 
Thus, in the inference stage, the LLMs can directly recite the causal knowledge without understanding the true causality in the context.
Similar behaviors are exhibited in a Causal Reasoning Assessment Benchmark, CRAB 
\cite{romanou2023crab}.

\citet{kim2023can} examined LLMs' abilities to understand the causalities of both scientific papers and newspapers.
The results show that ChatGPT performs worse than a fine-tuned BERT model in determining the causality or correlation of given statements.
\citet{abdali2023extracting} show the effectiveness of applying LLMs to diagnose the cause of issues from Microsoft Windows Feedback Hub.
\citet{li2023relation} showed that LLMs can identify dynamical (spatio-temporal) effects. However, how to infer the relationship and interactions of them is still challenging for LLMs, which are more emphasized as causal structures in causal inference.

Another important line of work is to understand LLMs' hallucination and faithfulness in knowledge reasoning by considering causal effects.
\citet{tang2023towards} proposed a multi-agent system, CaCo-CoT, where some LLMs are \textit{reasoners} and others are \textit{evaluators}.
\textit{Reasoners} try to provide causal solutions, while \textit{evaluators} try to challenge the \textit{reasoners} with counterfactual candidates.
With the cooperative reasoning framework, CaCo-CoT helps to improve 
causal-consistency.
\citet{zhang2023mitigating} identified the potential knowledge bias pretrained in the LLMs as the confounder which causes incorrect answers and hallucinations.
They proposed a CoT framework to generate sub-questions necessary to answer a question and require humans to provide the correct answers.
Moreover, \citet{chen2024causal} introduced CaLM, a comprehensive benchmark that systematically assesses LLMs’ causal reasoning across diverse tasks, adaptations, and error types.

\subsubsection{Commonsense Reasoning}
\label{subsec:commonsense}

Commonsense reasoning involves the ability to apply everyday knowledge and intuitive understandings of the world to make decisions or draw conclusions, which is vital for LLMs' contextual understanding and human-like interactions \cite{davis2015commonsense, storks2019commonsense}.
This section briefly summarizes the commonsense reasoning ability of LLMs under various settings \cite{gao2023chatgpt, willig2022probing} and the employment of causally motivated methods in improving commonsense causality reasoning \cite{zhang2022rock,zheng2023preserving,lu2022neuro,chen2023learning,zhao2023competeai}.

The reasoning ability of LLMs is limited but it could generate good causal explanations \cite{gao2023chatgpt}. This viewpoint has been validated through ChatGPT on event causality identification, causal discovery, and causal explanation generation.  \citet{willig2022probing} showed a similar observation by evaluating LLMs on causal question answering, where LLMs arrived at their answers through memorization instead of reasoning.

To improve the commonsense causality reasoning of LLMs that identify causes from effects in natural language, ROCK \cite{zhang2022rock} balances confounding effects using temporal propensities through an estimation of the average treatment effect. 
While ROCK adopts a potential outcome framework, \citet{chen2023learning} uses a conversation cognitive model based on intuition theories and transforms intuitive reasoning into a structural causal model. 

Other than facilitating the reasoning ability of LLMs directly, \citet{zheng2023preserving} use causal inference to preserve commonsense knowledge from pre-trained language models for fine-tuning to prevent catastrophic forgetting; 
\citet{lu2022neuro} focus on improving LLM’s ability in generalized procedural planning with commonsense-infused prompts. For procedural planning tasks, \citet{lu2022neuro} proposed to learn cause\-effect relations among complex goals and stepwise tasks, and reduced spurious correlation among them via front door adjustment.

\subsubsection{Counterfactual Reasoning}
\label{subsec:counterfactual}

LLMs can generate counterfactuals as data augmentations for small language models. Given a text $x$ and a black-box classifier $B$, the counterfactual text $\tilde{x}$ of text $x$ should satisfy: \cite{betti2023relevance, miao2023generating}: (1) $\tilde{x}$ has a different class than $x$, $B(x) \neq B(\tilde{x})$; (2) $x$ and $\tilde{x}$ differ only by minimal lexical changes; (3) $\tilde{x}$ is a feasible text and the commonsense constraint is satisfied.

To understand LLM's ability to generate counterfactuals, \citet{li2023large} examined the following tasks: sentiment analysis 
, natural language inference (NLI) 
, named entity recognition (NER)
, and relation extraction (RE) 
For simple tasks like sentiment analysis and NLI, data augmented via LLMs can mitigate potential spurious correlations. For more complicated tasks like RE, LLMs may generate low-quality counterfactuals. 
\citet{liu2023magic} evaluated abductive reasoning and counterfactual reasoning abilities and found code large language models (Code-LLMs) achieved better results compared to text models. 

Many research also showed different ways of incorporating counterfactuals in domain-specific tasks.
\citet{miao2023generating} claimed that existing methods for generating counterfactuals for RE face two challenges: identifying causal terms correctly and ignoring the commonsense constraint. To amend this, they proposed an intervention-based strategy to generate commonsense counterfactuals for stable relation extraction.
Similarly, \citet{oba2023contextual} and \citet{sen2023people} used counterfactual generation to address issues in gender bias and harmful language detection, respectively.


\subsection{Fairness and Bias}

\label{sec:fairness}
Fairness and bias are pivotal factors in deploying language models effectively and ethically. Bias is common in pretrained language models as they capture and potentially amplify undesired social stereotypes and biases \cite{meade2021empirical, achiam2023gpt, zhou2023explore, wu2024safety,wang2024mementos,gallegos2024bias}. An example of bias in language models includes gender associations with specific professions, such as male firefighters and female nurses \cite{stanovsky-etal-2019-evaluating}. Causality-based methodologies offer a promising approach for mitigating biases in language models by discerning the origins of bias through a causal perspective. Bias mitigation is then followed by eliminating the unwanted spurious correlation between generative factors through different types of causal intervention or causal invariant learning.

\citet{ding2022word} introduced a proxy variable related to gender bias in the causal graph, and used two different ways to eliminate the potential proxy bias and unresolved bias under the linear structural equation model. 
\citet{zhou2023causal} believe that the backdoor path between the ground truth label and the non-causal factors is the source of bias, and used the Independent Causal Mechanism principle to mitigate bias. 
\citet{wang2023causal}, from a different angle, eliminated the bias by performing a do operation on the intermediate variables for both white-box and black-box LLMs. 
\citet{madhavan2023cfl} considered the tokens generated by generative language models trained with causal language modeling objectives as a causal graph, and analyzed the bias under this model.
\citet{xia2024aligning} used a reward model as an instrumental variable to perform causal interventions to reduce biases in LLM outputs by modeling the confounders in pretraining data and input prompts.
\citet{wu2024decot} used external knowledge as an instrumental variable and front-door adjustment to improve the accuracy of CoT reasoning on knowledge-intensive tasks.
\citet{jenny2023navigating} used Activity Dependency Networks to describe the causality effect between normative variables, such as clarity and authenticity, to structure the cause of bias.

\subsection{Safety}
\label{sec:safety}

As researchers have observed the unreliability phenomenon of LLMs in various applications \cite{cao2022can, zhao2023causality, zhao2022certified, meng2021self, liu2023hallusionbench, xu-etal-2024-promises}, there is increasing interest in applying causal inference techniques to analyze the causality of the non-robustness of the model and adjust the treatment to resolve the challenges \cite{cao2022can, zhao2022certified, zhao2023causality}. 

LLMs face challenges of unreliability when performing knowledge probing by query LLMs with task-specific prompts \cite{cao2022can, xiao2023large}, such as using shortcuts to complete the probing and generating different predictions for the semantically equivalent prompt. 
By simply translating text tokens in input prompts to emoji sequences, LLMs generated more severe hallucinations \cite{zhao2023causality}.
The main reason is that they rely on spurious correlations to make an inference \cite{zhao2022certified}. Training LLMs to learn the causal relationship between input $x$ and output $y$ is an intuitive method of better resisting prompt attacks. The randomized smoothing technique \cite{zhai2020macer, jia2019certified} can model the interventional distribution $p(y | do(x))$ by assuming discrete adversarial perturbations as the Gaussian distribution \cite{zhao2022certified}. The method of smoothing in the latent semantic space is more robust against known attacks such as word substitutions, paraphrasing, and token position change \cite{zhao2022certified}.

\subsection{Explainability}
\label{sec:explanation}

Explainability in LLMs refers to the capacity to elucidate how these models arrive at their conclusions, enhancing transparency and trustworthiness in AI decision-making processes \cite{guidotti2018survey, lipton2018mythos}.
Many works have tried to understand the inner workings of LLMs~\cite{hou2023towards, belrose2023eliciting, gurnee2023finding, zhao2023explainability,liu2023aligning,li2023towards}. 
We summarize research efforts that probe the causal mechanism in LLMs from the following three directions: intervening the inputs or prompts, intervening inner components of LLMs, and abstracting the working mechanism into a causal graph.

\paragraph{Inputs or Prompt Intervention}
Input intervention, as a data-centric method, is to create counterfactual input text by changing the treated feature in the text, and then observe the model behaviors on the original and counterfactual texts. 
LLMs can first identify the features in input texts causally associated with the predictions and are capable of changing the identified features to create the counterfactual texts \cite{bhattacharjee2023llms, gat2023faithful}. These counterfactual texts can be utilized to investigate the causality of the LLM and can serve as a training dataset to learn a matching model, where the matched counterfactual pairs have similar embeddings \cite{gat2023faithful}.

Various works have developed different prompting methods and found whether the prompting methods are causally associated with the final output of LLMs \cite{schick2020exploiting, zhou2023scalable, wei2022chain}. However, the causal effect of prompting methods, such as chain-of-thought (CoT), on the final output is ambiguous. Prompt intervention, which alters only one particular aspect of prompts, is proposed to understand the contributions of each component of prompts on model behavior \cite{madaan2023makes, ji2023benchmarking,tan2023causal}. 
These findings suggest that LLMs rely on the causal model suggested by their CoTs to a high extent, but LLMs also learn spurious correlations such as sentence length to generate responses.

\paragraph{Inner Component Intervention}
Input or prompt interventions help understand model behavior, but inner component intervention is needed to uncover the information cascade within LLMs. Existing works mainly focus on the essential components in SoTA LLMs: attention mechanism and MLP.
\citet{stolfo2023mechanistic} exchanged the activation values in MLP and attention layers of different inputs to probe the function of MLPs and attention mechanism.
While \citet{stolfo2023mechanistic} focuses only on math word problems with four fundamental arithmetic operators, it is an interesting direction to generalize the component intervention to other applications. 
\citet{vig2020investigating} used causal mediation analysis to interpret which components of neural models are causally implicated in their behaviors and applies this approach to uncover how gender bias is mediated in GPT2.

\paragraph{Causal Graph Abstraction}
An intuitive way to characterize causality within LLMs is to abstract the working mechanism of LLMs into a causal graph. 
Boundless Distributed Alignment Search (DAS) \cite{wu2023interpretability}, by replacing brute-force searching the original DAS \cite{geiger2023finding} with learnable parameters, 
has been effective on the Alpaca model \cite{taori2023alpaca}. Given four pre-defined causal models, Boundless DAS extracts two of them as the accurate hypotheses as the abstracted causal graph of the Alpaca model. However, the Boundless DAS method is restricted by the given causal hypothesis, and the future direction can explore how to abstract the causal graph in LLMs without prior causal graphs.

\subsection{Multi-modality} 
\label{sec:multimodal}
Large vision-language models have been popular in many applications~\cite{maaz2023video,li2023blip,liu2023visual, guan2024loczson}. 
How to conduct causal reasoning on both images and texts can be crucial in correctly answer multimodal questions \cite{niu2021counterfactual}.
\citet{pawlowski2023answering} examined LLMs' causal reasoning abilities and showed that the causal knowledge in the language models can be too strong a prior which often causes the model to neglect visual evidence. 
\citet{si2022language} highlighted how VQA models tend to rely on various shortcuts beyond language priors.
\citet{ko2023large} proposed to alleviate the problem by adding self-consistent generation prediction by checking predictions based on different parts of the input, in which the three inputs V, Q, and A are individually predicted based on the other two inputs.
Additionally, \citet{li2023image} proposed an image generation framework with causal reasoning and created a novel VQA dataset whose questions require causal explanations. \citet{chen-etal-2024-quantifying} curated a multi-hop VQA dataset, assessed unimodal biases in LVLMs, and proposed a causality-enhanced agent framework to mitigate them.

Another important question is to understand the spatial-temporal causal relationship of the visual elements within the images and videos. \citet{su2023language} proposed CaKe-LM to use pretrained causal knowledge for video understanding while \citet{tai2023link, zhao2023causal} designed specific structure and prompting method to capture and interpret the underlying causal relationship for VQA.
\subsection{Evaluation and Benchmark}
\label{sec:eval}
In this section, we list existing LLM benchmarks built from a causal perspective.




\begin{table}[h]
\small
\centering
\label{tab:benchmark-and-eval}
\resizebox{\columnwidth}{!}{%
\begin{tabular}{l|cccc|cc}
    \toprule
    Reference                                       & \textbf{MU}   & \textbf{CR}   & \textbf{CF}    & \textbf{FD}    & \textbf{Text}  & \textbf{MM} \\ \midrule
    ECHo \cite{xie2023echo}                         & \checkmark    & \checkmark    &                &                &                     & \checkmark          \\
    CREPE \cite{zhang2023causal}                    &               &  \checkmark   &               &                 & \checkmark          &            \\
    CLOMO \cite{huang2023clomo}                     &               &               & \checkmark     &                & \checkmark          &            \\
    IfQA \cite{yu2023ifqa}                          &               &               & \checkmark    &                 & \checkmark          &            \\
    Cladder \cite{jin2023cladder}                   & \checkmark    & \checkmark    &                &                & \checkmark          &            \\
    MoCa \cite{nie2023moca}                         & \checkmark    &               &                &                & \checkmark          &            \\
     \citet{jin2023can}                    & \checkmark    &               &                &                & \checkmark          &            \\
    CVidQA \cite{su2023language}                    & \checkmark    &               &                &                &                     & \checkmark          \\
    VQAI \cite{li2023image}                         & \checkmark    &               &                &                &                     & \checkmark          \\ 
    \citet{chen2023models}               &               &               & \checkmark     &                & \checkmark          &            \\
    \citet{gao2023chatgpt}                & \checkmark    &               &                &                & \checkmark          &            \\ 
    CRAB \cite{romanou2023crab}                     & \checkmark    &               &                &                & \checkmark          &            \\
    HELM \cite{liang2022holistic}                   &               &               &                &    \checkmark    & \checkmark          &            \\
    Fair-Prism \cite{fleisig2023fair}               &               &               &                &    \checkmark    & \checkmark          &            \\ 
    Biasasker \cite{wan2023biasasker}               &               &               &                &    \checkmark    & \checkmark          &            \\ 
    CaLM \cite{chen2024causal}            &      \checkmark         &       \checkmark        &      \checkmark          &       & \checkmark          &            \\ 
    MORE \cite{chen-etal-2024-quantifying}   & \checkmark  &     &     &    &   & \checkmark \\
    \bottomrule
\end{tabular}%
}
\caption{Existing evaluation benchmarks. Based on the evaluation tasks, we categorize the benchmarks into three categories: Model Understanding (\textbf{MU}), Commonsense Reasoning (\textbf{CR}), Counterfactual Reasoning (\textbf{CF}) and Fairness/Debias (\textbf{FD}). Based on the modalities of the data samples, we identify the benchmarks with only textual inputs (\textbf{Text}) and those with multimodal inputs (\textbf{MM}).}
\label{tab:evaluate}
\end{table}

The benchmarks in model understanding (MU) focus on evaluating and understanding LLMs' causal reasoning abilities in both 
natural language \cite{jin2023cladder,nie2023moca,jin2023can,gao2023chatgpt,romanou2023crab, chen2024causal} and vison-language \cite{su2023language,li2023image,xie2023echo, chen-etal-2024-quantifying}.
In addition, some benchmarks \cite{nie2023moca,gao2023chatgpt} also provide model understanding in comparison with human causal reasoning and moral judgments.
Commonsense reasoning benchmarks (CR) evaluate LLMs on tasks that require extensive commonsense knowledge for both textual-only context \cite{zhang2023causal,jin2023cladder} and multimodal context \cite{xie2023echo}.
Contexts with commonsensical and anti-commonsensical are constructed in \cite{jin2023cladder}, to further investigate whether LLMs use averaged-out causal reasoning.
Evaluating LLMs' counterfactual reasoning (CF) abilities is essential in enabling explainable model reasoning and calibration of the generated rationales.
\citet{huang2023clomo} introduce a specific task and benchmark for assessing LLMs' logical counterfactual thinking. 
\citet{yu2023ifqa} contribute a novel dataset to challenge LLMs in counterfactual reasoning in an open-domain QA context.
\citet{chen2023models} investigate the ability of LLMs to provide explanations that aid in understanding their reasoning process, particularly in the context of counterfactual scenarios.
The fairness and bias (FD) evaluations are particularly in addressing biases, fairness, and the overall transparency of language models.
HELM \cite{liang2022holistic} is a comprehensive evaluation benchmark including previously neglected areas for fairness.
Fair-Prism \cite{fleisig2023fair} focuses specifically on fairness-related harms in models, which are identified and measured by detailed human annotations.
Biasasker \cite{wan2023biasasker} presents an automated framework to identify and measure social biases by probing the models with specially designed questions.


\section{Large Language Model for Causal Inference}

Causal inference, as a powerful tool for developing large language models (LLMs), greatly benefits from the extensive world knowledge that LLMs provide. We summarize how LLMs can help causal inference in its two important components, i.e., causal relationship discovery and treatment effect estimation. 

\subsection{Treatment Effect Estimation}
\label{subsec:treatment}
Estimating treatment effects is central to causal inference but is hindered by the absence of counterfactual data in many cases. 
\citet{chen2023disco} proposed a new method for automatically generating high-quality counterfactual data at scale called DISCO (DIStilled COunterfactual Data). In addition, ~\citet{feder2023causal} apply treatment effect estimation to align knowledge for generalization towards different domains.
\citet{zhang2023towards} optimize the treatment effect estimation on unlabeled datasets by performing self-supervised causal learning through LLMs. 

\subsection{Causal Relationships Discovery}
\label{subsec:discovery}
Discovering causal relationships between variables is fundamental in causal inference as it enables the identification and estimation of causal effects. Many research focus on casual relationship extraction or causality extraction which extracts causal relationships between two variables from text directly. Traditional methods rely on linguistic cues such as causal connectives (e.g., ``cause'', ``because'', and ``lead to'') and grammatical patterns to identify causal pairs \cite{xu2020review}. A later work utilizes the power of statistical machine learning and deep learning to tackle this task in a supervised learning setting \cite{yang2022survey}. 
As LLMs show promising potential with reasoning capacities as introduced in Section \ref{sec:reasoning_capacity}, many works use LLMs as a query tool to determine the edge direction between two given variables. For example, \citet{kiciman2023causal} show that LLMs can achieve competitive performance in determining such pairwise causal relationships with accuracies up to $97\%$. 
Analyses in the medical domain \cite{naik2023applying, Antonucci2023ZeroshotCG, arsenyan2023large} exhibit similar observations. However, other studies highlight LLMs' limitations of such pairwise causal relationships. For example, sensitivity to prompt design leads to inconsistent results \cite{long2023can}; pairwise judgments can lead to cycles in the full causal graph \cite{vashishtha2023causal}; pairwise judgments require large computational cost when applying to a large-scale dataset, $N$ variables would require $\binom{N}{2}$ prompts \cite{ban2023causal}; LLMs still provide false information despite achieving strong results in many cases \cite{long2023can, tu2023causal, joshi2024llms}. \citet{long2023can} propose strategies for amending LLMs' output based on consistency properties in causal inference.
\citet{feng2024pre} found the LLMs' performance on causal discovery tasks depends on pretraining corpora (corpora with higher frequency of causal mentions perform better) and provided context when making the predictions.

To alleviate the impact of erroneous causal information from LLMs, previous works have integrated LLMs with traditional causal discovery methods. Causal discovery or causal structure learning is the task of recovering causal graphs from observational data whenever possible \cite{zanga2022survey}.\citet{vashishtha2023causal} propose two algorithms that combine LLMs with causal discovery methods: the first uses causal order from an LLM to orient the undirected edges outputted by a constraint-based algorithm and the second utilizes the LLM causal order as a prior for a score-based algorithm. 
Similarly, \citet{li2024realtcd} introduced LLM-guided meta-initialization to extract the meta-knowledge from textual information to improve the quality of a temporal causal discovery method.
\citet{khatibi2024alcm} used LLMs to refine the output of data-driven causal discovery algorithms.
\citet{ban2023query} incorporated LLM-driven causal statements as qualitative ancestral constraints in the Bayesian network structure to guide data-driven algorithms. \citet{ban2023causal} then propose an iterative framework to validate and finetune based on LLM feedback.
For a broader discussion on integrating LLMs into causal discovery methods, we refer readers to a recent survey on this specific topic \cite{wan2024bridging}.

\subsection{Future Directions}
LLMs can significantly contribute to overcoming the current limitations of causal inference methods as a general expert. A common assumption in many causal methods is the existence of corresponding data points for every treatment. However, this assumption often proves untrue, particularly when dealing with imbalanced minority data that may not support meaningful learning. LLMs, functioning as versatile experts, have the potential to address this challenge by aiding in data augmentation for minority data. Through their comprehensive understanding of language and context, LLMs can enhance the availability of diverse data points, facilitating more robust and effective causal inference in situations where traditional methods may struggle due to data imbalances. Similarly, many methods operate under a strong assumption of unconfoundedness within the potential outcome framework
Historically, this assumption has been accepted due to a lack of domain knowledge regarding the underlying causal graph or identification of potential confounders. However, LLMs offer an opportunity to alleviate the limitation.


\section{Conclusion}
\label{sec:conclusion}

At its core, a large language model (LLM) is like a vast library of knowledge. One of the ongoing challenges is figuring out how to extract and use this knowledge effectively. The key to improving LLMs lies in enhancing their ability to understand cause and effect – essentially, how things are connected. Causal reasoning is crucial for making LLMs smarter. Looking at it from a causal inference perspective, we find a valuable framework that helps boost the effectiveness of LLMs. Meanwhile, as keepers of human knowledge, LLMs can even help overcome limitations in causal inference by providing broad expertise that goes beyond existing constraints, reshaping our understanding in this important area and bringing new vitality to this area. In this survey, we offer a thorough examination of both directions and concise summaries of the methods scrutinized, offering a comprehensive overview of the current state of research at this intersection. 

\section*{Limitations}
\label{sec:limitations}
With the increasing popularity of large language models (LLMs), understanding their reasoning abilities becomes ever more crucial. Many tasks performed by LLMs require an understanding of causality, making the evaluation and enhancement of their reasoning and causal inference abilities a key focus. This is where causal inference plays a critical role. At the same time, causal inference itself relies on a certain level of world knowledge, which LLMs are well-suited to provide. In this survey paper, we aim to provide a comprehensive review of how LLMs contribute to causal inference and, in turn, how causal inference can improve LLMs. 
However, the intersection of causal inference and LLMs represents a rapidly evolving research frontier where significant developments appear across NLP/ML venues and preprints. 
Given this field’s dynamic nature, our selection methodology prioritizes comprehensive coverage of key developments over exhaustive enumeration. For preprints, the authors of this paper manually reviewed them to assess their quality and relevance to the topic.
There are several possible ways to organize this paper, particularly for Section \ref{sec:causal_for_llm}. We follow the framework proposed by \citet{feder-etal-2022-causal} on how causal inference can enhance NLP models, focusing on performance, robustness, fairness, and interpretability. While \citet{feder-etal-2022-causal} focuses on embedding-based methods, our survey reflects the transformative impact of LLMs with strong generative capabilities. We supplement the categorization with additional discussions on LLM’s reasoning capacity and multi-modality, capturing research trends driven by the enhanced capacities of LLMs. 
It’s worth noting that some papers may fit into multiple sections, and we grouped them based on our judgment of the papers' main contributions.


\bibliography{reference}

\clearpage

\appendix

\section{Brief Introduction of Causal Inference}
\label{sec:causal-basics}

In this section, we present the background knowledge of causal inference, including task descriptions, basic concepts and notations, and general solutions.

Generally speaking, the task of causal inference is to estimate the causal relationship among variables. The variables of interest are referred to as \textit{treatment}, naturally, the effects of treatments are referred to as \textit{treatment effects}. 
For example, suppose two treatments can be applied to patients: Treatment Plan A and B. When A is applied to a certain patient cohort, the recovery rate is $70\%$ while when B is applied to \textit{the same cohort}, the recovery rate is $80\%$. The change of recovery rate is the effect of that treatment assets on the recovery rate. 

Ideally, the treatment effect can be measured as follows: applying different treatments to the same cohort, and then the difference in the effect is the treatment effect. However, in real-world scenarios, this ideal situation because it is impracticable for perfectly controlled experiments in most cases. For example, in the above case, you can only apply one treatment to the same cohort at the same time. In reality, an alternative is to conduct random controlled trials, in which the treatment assignment is controlled, such as a completely random assignment. In this
way, the groups receiving different treatments can be used to measure the difference in effect. Unfortunately, even performing randomized experiments is expensive, time-consuming, and may cause ethical concerns in some cases.
Therefore, estimating the treatment effect from observational data has attracted growing attention due to the wide availability of observational data, and methods are developed for the investigation of the causal effect of a certain treatment without performing randomized experiments. %


\subsection{Potential Outcome Framework}
One of the most influential frameworks in identifying and quantifying causal effects in observational data is the potential outcomes framework \cite{rubin1974estimating}. The potential outcomes approach associates causality with manipulation applied to \textit{units}, and compares causal effects of different treatments via their corresponding potential outcomes. Following \cite{rubin1974estimating}, we state basic concepts in the potential outcome framework.

\noindent\textbf{Unit.} A unit is the atomic research object in the treatment effect study. A unit can be a physical object, a firm, a patient, a person, or a collection of objects or persons, such as a classroom or a market, at a particular time point \cite{rubin1974estimating}. Under the potential outcome framework, the atomic research objects at different time points are different units. 


\noindent\textbf{Treatment.}  Treatment refers to the action that applies (exposes, or subjects) to a unit. For each unit-treatment pair, the outcome of that treatment when applied to that unit is the \textbf{potential outcome}.With N treatments $T = \{1,2,3,..., N\}$, the potential outcome of applying treatment $T_i$ is denoted as $Y(T=T_i)$. The \textbf{observed outcome} is the outcome of the treatment that is actually applied. And the \textbf{counterfactual outcome} is the outcome if the unit had taken another treatment.





\noindent\textbf{Treatment Effect} The treatment effect can be quantitatively defined using the above definitions. The treatment effect can be measured at the population, treated group, subgroup, and individual levels. At the population level, the treatment effect is estimated as the Average Treatment Effect (ATE). At the subgroup level, the treatment effect is called the Conditional Average Treatment Effect (CATE).

\begin{definition}[Binary Average Treatment Effect(ATE)] Suppose we want to measure the treatment effect of a treatment $T=1$. Then the average treatment effect is defined as:
\begin{equation}
    \label{def:ate}
    \mathbb{E}[Y(T=1) - Y(T=0)]
\end{equation}
where $Y(T=1)$ and $Y(T=0)$ denote the potential treated and control outcome of the whole population respectively.
\end{definition}

\begin{definition}[Conditional Average Treatment Effect (CATE)]
    \label{def:cate}
    \begin{equation}
        \mathbb{E}[Y(T=1)|X=x] - \mathbb{E}[Y(T=0)|X=x]
    \end{equation}
where $\mathbb{E}[Y(T=1)|X=x]$, $\mathbb{E}[Y(T=0)|X=x]$ are the potential treated and control outcome of the subgroup with $X = x$.
\end{definition}

At the individual level, the treatment effect is defined as Individual Treatment Effect (ITE). In some literature, ITE is treated as the same as CATE~\cite{pearl2009causality}.



\subsection{Causal Graphical Models}
The potential outcome framework is powerful in recovering the effect of causes. In a potential outcome framework, causal effects are answered by specific manipulation of treatments. However, when it comes to identifying the causal pathway or visualizing causal networks, the potential outcome model has its limitations. In the front of the challenge, causal graphical models utilize directed edges to represent causalities and encode conditional independence among variables in the graphs.

\subsubsection{Structural Equation Models (SEMs)}
One of the most widely-spread formulations is the Structural Equation Model \cite{wright1934method, pearl1998graphical}, where linear structural equation models are used to present causal relationships by directed edges, which differentiate correlation from causation when the graph structure is given. The linearity assumption was later been relaxed by \cite{pearl1998graphical} and it formalized causal graphical models for presenting causal relations using Directed Acyclic Graphs (DAGs).

Specifically, consider the random variable $\mathbf{X} \in \mathcal{R}^{D \times N} = [X_1, X_2, ..., X_N]$,  the linear SEM consists of a set of equations of the form:
\begin{equation}
    \label{def:linear-sem}
    X_i = \beta_{0i} + \sum_{j \in pa(X_i)} \beta_{ji}X_j + \epsilon_i, \quad i = 1,2,3,..., N
\end{equation}
where $pa(X_i)$ denotes the set of variables that are direct parents of $X_i$. $\epsilon_1, \epsilon_2, ..., \epsilon_N$ are mutually independent noise terms with zero mean, $\beta_{ji}$ are coefficients that quantify the causal effect of $X_j$ on $X_i$.

While the non-parametric SEM takes the form:
\begin{equation}
    \label{def:non-linear-sem}
    X_i = f_i(\mathbf{X}_{pa(i)}, \epsilon_i), \quad i = 1,2,3,..., N
\end{equation}

The random variables $\mathbf{X}$ that satisfies the model structure of the form in \Cref{def:linear-sem} or \Cref{def:non-linear-sem} can be represented by a directed acyclic graph (DAG) $G = (V, E)$, where $V$ is the set of associated vertices, each corresponding to one of a variable of interest $X_i$, and $E$ is the corresponding edge set.

With pre-specified DAG and assumptions on the latent variables, the coefficients between the latent variables are identifiable\cite{kuroki2014measurement}.

\subsubsection{Bayesian Network}
Causal inference can be naturally embedded in graphical model frameworks since the dependencies and interactions between variables can be presented by graphs with probabilistic distributions, in which nodes correspond to variables of interest and edges represent associations. One general solution except for SEMs is to use a Bayesian Network to represent the causal relationship. 

In Bayesian networks, causalities among variables are represented in the form of DAGs with directed edges carrying causal information.

A joint probability distribution $\mathbb{P}$ factorizes with respect to a DAG $\mathcal{G}$ if it satisfies:
\begin{equation}
    f(X_1, X_2, ..., X_N) = \prod_if(X_i|\mathbf{X}_{pa(i)})
\end{equation}

In the next section, we show a comprehensive survey of how existing works help with the tasks and challenges in LLMs in detail.






\end{document}